\documentclass[conference]{IEEEtran}
\usepackage{cite}
\usepackage{amsmath,amssymb,amsfonts}
\usepackage{algorithmic}
\usepackage{graphicx}
\usepackage{textcomp}
\usepackage{xcolor}
\usepackage{booktabs}
\usepackage{array}
\usepackage{makecell}
\usepackage[T1]{fontenc}
\usepackage[utf8]{inputenc}
\usepackage{microtype}
\usepackage{url}
\usepackage{tikz}
\usetikzlibrary{arrows.meta,positioning}

\definecolor{promptbg}{gray}{0.96}
\definecolor{promptborder}{gray}{0.80}

\begin{document}

\title{TRACE-BN: Transferring Bangla-English Tutoring Behavior to a Sub-1B Offline Language Model}

\author{
\IEEEauthorblockN{Khan Raiyan Ibne Reza, Sanjana Aktar Maria, Mohammad Tushar Abdullah,\\Asfee Bhuiyan Leen, and Sumaiya Tabassum Nimi}
\IEEEauthorblockA{Department of Electrical and Computer Engineering, North South University, Dhaka, Bangladesh\\
Email: \{raiyan.reza, sanjana.maria, tushar.abdullah.232, asfee.leen.242, sumaiya.nimi\}@northsouth.edu}
}

\maketitle

\begin{abstract}
Bangla-English tutoring requires more than producing a correct translation: learners also need explanations of grammar differences, awareness of their likely errors, and targeted practice. We present \textbf{TRACE-BN}, a curriculum-guided dataset of structured tutoring traces for Bangla-speaking learners of English at the CEFR A1--A2 level. Each trace combines word-level glosses, literal and natural translations, Bangla grammar explanations, a plausible learner error, and a targeted practice question with its answer. The traces are generated by Gemini 3.5 Flash Lite as the teacher model from NCTB Classes 9--10 English curriculum units, then filtered for structural validity, script integrity, and semantic duplication. We transfer the resulting structured tutoring behavior to Qwen3-0.6B using LoRA with 4-bit quantization for resource-constrained offline deployment. On held-out inputs, schema validity increases from 85.4\% to 95.8\%, while, against teacher-model references, chrF++ improves from 15.28 to 34.77 and BLEU from 4.52 to 21.03. Field-level evaluation by two independent judges shows improvements across translation, grammar explanation, learner-error diagnosis, and practice alignment, while a human audit supports the quality of the supervision data. The results show that curriculum-guided structured supervision can transfer multi-component tutoring behavior to a sub-1B model under these resource constraints. The dataset, model checkpoints, and code are publicly available at \url{https://huggingface.co/datasets/RaiyanKhaan/Trace-BN}.
\end{abstract}


\section{Introduction}

A Bangla-English tutoring system should do more than produce a target translation: it should explain relevant grammatical differences, identify likely learner errors, and provide practice on the same pattern. Large language models can support this interaction more fully than conventional translation systems can, but existing Bangla-focused models are not designed around a structured tutoring interaction. Such an interaction requires word-level glosses, literal and natural translations, Bangla grammar explanations, prediction of likely learner errors, and targeted practice within a single response. Existing Bangla models such as TigerLLM, BanglaLlama, and TituLLM provide general language capabilities, while recent low-resource tutoring systems include substantially larger models and broader multimodal settings~\cite{raihan2025tigerllm,zehady2026banglallama,kabir2025titullms,mahfuz2025too,belay2026afrilangtutor}.

We address this problem with \textbf{TRACE-BN}, a curriculum-guided dataset of 4,099 structured Bangla-to-English tutoring traces. Each trace contains seven fields: word-level glosses, literal and natural translations, Bangla grammar notes, a common learner mistake, a practice question, and its answer. Unlike a conventional parallel corpus, each TRACE-BN instance encodes a complete instructional sequence rather than a single input-output pair. Learner-facing scaffolding is provided in Bangla, while target-language content remains in English. The design draws on second-language acquisition research emphasizing contrastive explanation and structured practice~\cite{ellis2006implicit,vanpatten2004processing,cook2010translation,leonardi2010role}.

We then investigate whether these traces can teach the underlying tutoring behavior to a sub-1B model. We fine-tune Qwen3-0.6B with LoRA and evaluate it against the untuned base model and three zero-shot baselines. On 432 held-out examples, the tuned model improves schema validity from 85.4\% to 95.8\% and raises chrF++ from 15.28 to 34.77 and BLEU from 4.52 to 21.03. A field-level evaluation with two independent judges further shows consistent improvements across translation, grammar explanation, learner-error diagnosis, and practice alignment.

Our contributions are:
\begin{itemize}
\item \textbf{TRACE-BN}, a curriculum-guided bilingual tutoring dataset with structured traces across seven fields spanning translation, explanation, learner-error modeling, and practice.
\item \textbf{A structured tutoring task formulation} that combines translation with contrastive grammar explanation, learner-error prediction, and targeted practice in a single trace, extending beyond the Bangla-English pair studied here.
\item \textbf{An empirical adaptation study} showing that Qwen3-0.6B can learn to generate the TRACE-BN structure under LoRA fine-tuning, together with field-level evaluation and human validation of the supervision data.
\end{itemize}

\section{Related Work}
\label{sec:related_work}

\begin{figure*}[t]
\centering
\includegraphics[width=0.80\textwidth]{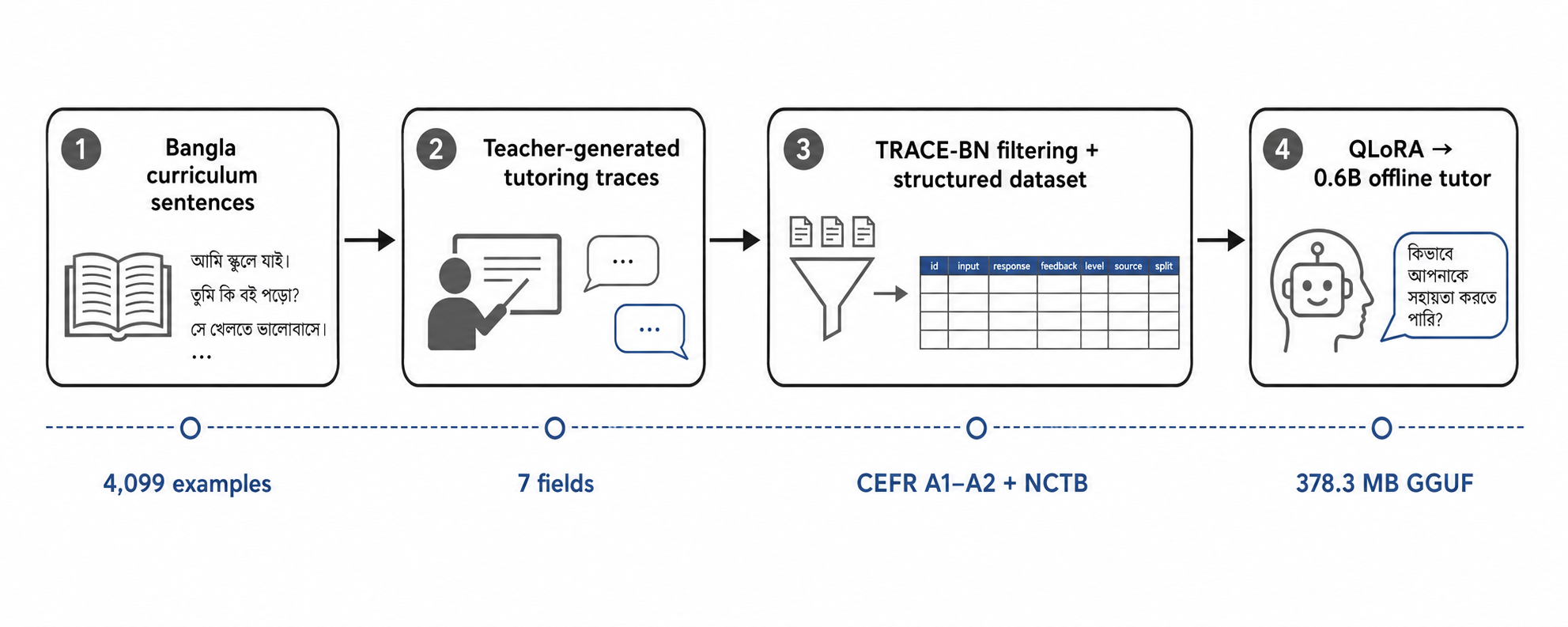}
\caption{TRACE-BN construction and adaptation pipeline. Curriculum-guided Bangla inputs are used to obtain teacher-generated tutoring traces, which are filtered into TRACE-BN and used to adapt Qwen3-0.6B with LoRA.}
\label{fig:pipeline}
\end{figure*}

\subsection{Bangla NLP and Low-Resource Language Tutoring}

Fig.~\ref{fig:pipeline} summarizes the TRACE-BN construction and adaptation pipeline. Recent Bangla language models, including TigerLLM, BanglaLlama, and TituLLM, target general Bangla language understanding and generation, while KrishokChat adapts a Bangla LLM to a single domain, agricultural advisory~\cite{raihan2025tigerllm,zehady2026banglallama,kabir2025titullms,reza2026krishokchat}. None of these systems combine translation with grammar explanation, error prediction, and practice generation in a single tutoring interaction.

Low-resource tutoring systems such as AfriLangTutor, LEARN, CaptainA, and LangLearn support AI-assisted language learning in other languages, but at a different scale and scope: AfriLangTutor fine-tunes 8B and 12B models on multi-turn dialogue across ten African languages, LEARN builds oral proficiency through cartoon-based visual question answering, CaptainA targets pronunciation practice, and LangLearn targets flashcard-based vocabulary drills~\cite{belay2026afrilangtutor,tushar2026personalized,phan2023captaina,zhang-etal-2025-learning}. TRACE-BN instead targets a sub-1B model with one schema spanning translation, grammar explanation, and error-aware practice for a single language pair.

A separate feasibility study concludes that Bangla LLMs are needed but that the field still lacks the high-quality pretraining and instruction-tuning data required to build them~\cite{mahfuz2025too}. TRACE-BN targets this gap for one downstream use: every trace is teacher-generated and then filtered for structural validity, script integrity, and semantic duplication before it is used for tuning (Section~\ref{sec:dataset}).

\subsection{Small Models and Structured Generation}

KD-LoRA, DistilQwen2.5, LoRA-Gen, and PhoneLM each specializes or deploys smaller language models through a different efficiency strategy~\cite{azimi2024kd,wang2025distilqwen2,xiao2025lora,yi2024phonelm}. TRACE-BN applies this direction to structured tutoring with a 0.6B target model.

Structured generation introduces a separate reliability issue: a response can satisfy a required schema without providing correct or useful content. Prior work has studied schema adherence and the effects of format constraints on language-model quality~\cite{pokrass2024introducing,lu2025learning,tam2024let,schall2025hidden,yang2025structeval}. Accordingly, our evaluation separates structural validity from translation and pedagogical quality.

\section{TRACE-BN Dataset Construction}
\label{sec:dataset}

\begin{figure*}[t]
\centering
\includegraphics[width=0.80\textwidth]{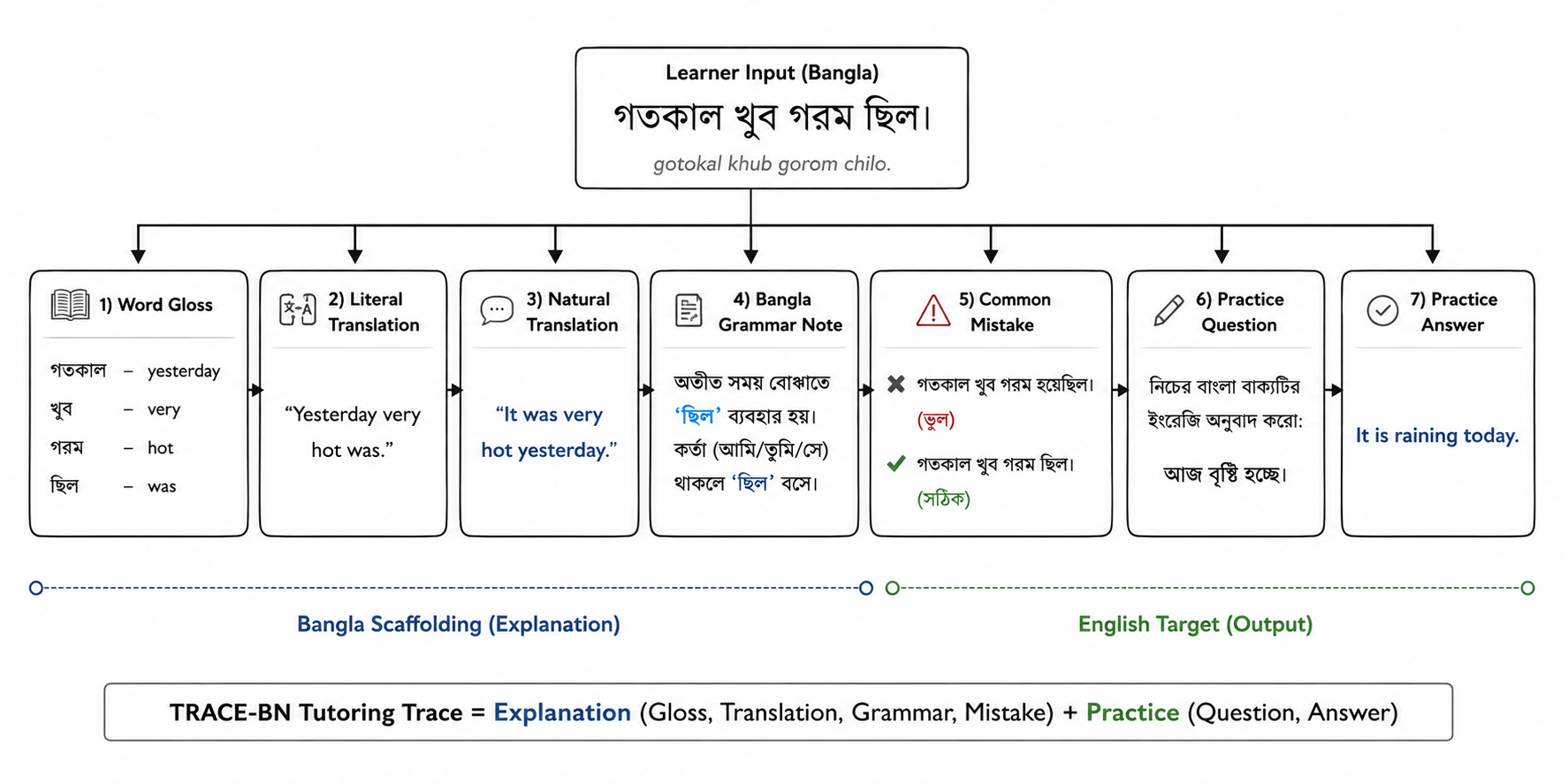}
\caption{Example TRACE-BN tutoring trace. A Bangla learner input is mapped to word-level glosses, literal and natural English, a Bangla contrastive explanation, a common learner error, and a short practice exercise with its answer.}
\label{fig:example}
\end{figure*}

TRACE-BN is based on the NCTB (National Curriculum and Textbook Board) English curriculum for Classes 9--10. We manually selected and reviewed the units, topics, and grammar coverage to define the dataset scope. Fig.~\ref{fig:example} shows a complete trace from the resulting dataset.

For each selected unit, we specified the target learner level, grammar or language pattern, and an everyday context. The specifications covered common Bangla--English differences relevant to beginners, including question formation, tense and auxiliary use, prepositions, introductory \textit{it} and \textit{there}, passive constructions, and idiomatic expressions.

We used these specifications to prompt Gemini 3.5 Flash Lite as the teacher model. Given the curriculum unit, target CEFR (Common European Framework of Reference) level, and Bangla input sentence, we instructed the model to generate a strictly valid JSON trace containing the seven fields shown in Fig.~\ref{fig:teacher_prompt}. The prompt also required no free-form text outside the JSON object.

\begin{figure}[t]
\centering
\fcolorbox{promptborder}{promptbg}{%
\begin{minipage}{0.95\columnwidth}
\fontsize{6.8pt}{8.0pt}\ttfamily
\textbf{[Teacher Generation Prompt Template]}\\
\textbf{System:} You are an expert English tutor for NCTB Class 9--10 exams.\\
\textbf{Input:} Unit: \textit{\{unit\}}, CEFR: \textit{\{cefr\}}, Bangla: "\textit{\{sentence\}}"\\
\textbf{Task:} Return strictly valid JSON conforming to:\\
\{\\
\hspace*{0.8em}"word\_gloss": [\{"bn": "..", "en": "..", "pos": ".."\}],\\
\hspace*{0.8em}"literal\_translation": "Word-for-word English",\\
\hspace*{0.8em}"natural\_translation": "Fluent English target",\\
\hspace*{0.8em}"grammar\_notes": ["Bangla L1 contrastive notes", ".."],\\
\hspace*{0.8em}"common\_mistake": "Plausible beginner L1 error",\\
\hspace*{0.8em}"practice\_question": "Targeted test question",\\
\hspace*{0.8em}"practice\_answer": "Expected correct answer"\\
\}\\
\textbf{Constraint:} Output NO text outside JSON. No markdown fences.
\end{minipage}%
}
\vspace{-2pt}
\caption{Teacher prompt template used for Gemini 3.5 Flash Lite trace generation.}
\label{fig:teacher_prompt}
\vspace{-4pt}
\end{figure}

\subsection{Curriculum Coverage and Language Design}

TRACE-BN contains 4,099 examples spanning 15 CEFR A1--A2 topic clusters and 13 NCTB grammar units. Topics include family, school, food, health, weather, directions, shopping, and daily routines. Grammar coverage includes passive voice, verbs and tenses, pronouns, prepositions, modals, tag questions, conditionals, sentence transformation, indirect narration, and introductory \textit{it} and \textit{there}.

Learner-facing scaffolding, including grammar notes and practice questions, is provided in Bangla, while target-language fields remain in English.

\subsection{Trace Filtering and Final Dataset}

We apply three automatic quality filters before training. First, structural validation requires valid JSON with all seven fields, non-empty grammar notes, and a word-gloss length between $0.5\times$ and $2.0\times$ the source sentence length. Second, Unicode script validation checks the Bangla source and target translation fields to prevent cross-lingual language-swapping errors. Third, semantic deduplication uses sentence embeddings from \texttt{paraphrase-multilingual-MiniLM-L12-v2} and removes near-duplicates with cosine similarity $>0.92$ within the same topic cluster.

From 4,450 teacher-generated candidates, 351 (7.9\%) were removed: 180 for structural errors, 45 for script-integrity violations, and 126 as semantic duplicates. The final dataset contains 4,099 traces, with mean Bangla source and English target lengths of 5.5 and 6.6 words, respectively.

The dataset is partitioned into 3,667 training examples and a 432-example evaluation split. The split is organized by topic clusters, including held-out domains such as weather conditions and idiomatic expressions, to test both sentence-level and topic-level generalization. No source-sentence overlap or detected lexical leakage exists between the evaluation and training sets.

As a human validation of the supervision signal, three bilingual English/Bangla language educators independently evaluated 100 randomly sampled TRACE-BN traces across the curriculum. Across five pedagogical dimensions on a 1--5 Likert scale, the supervision traces achieved mean ratings of $4.81 \pm 0.58$ for translation quality, $4.81 \pm 0.62$ for grammar explanations, $4.79 \pm 0.68$ for learner-mistake plausibility, $4.44 \pm 1.17$ for practice-question alignment, and $4.71 \pm 0.59$ overall ($95.7\%$ rated $\ge 4$). No factual or grammatical error reached annotator consensus in the audited sample. This human validation supports the quality of the supervision data, while the larger dual-judge evaluation assesses the behavior learned by the adapted model. Table~\ref{tab:dataset-summary} records the full dataset composition.

\begin{figure*}[t]
\centering
\includegraphics[width=0.80\textwidth]{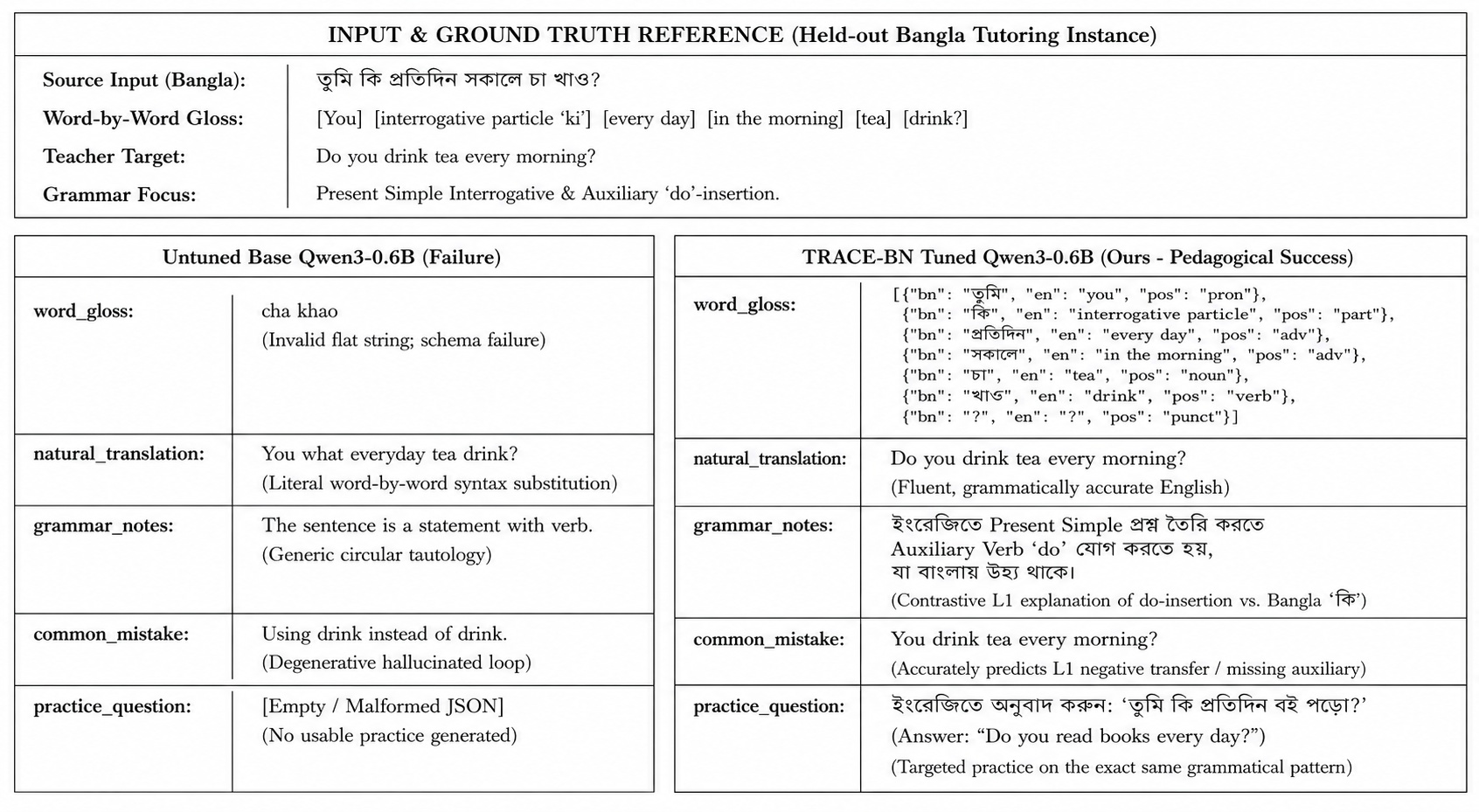}
\caption{Qualitative comparison of base vs. TRACE-BN-tuned Qwen3-0.6B outputs on representative held-out Bangla inputs. The base model produces literal word substitutions or degenerative loops, whereas the tuned model outputs fluent target translations alongside contrastive Bangla grammar explanations.}
\label{fig:qualitative}
\end{figure*}

\begin{table}[t]
\centering
\caption{Key specifications and composition of the TRACE-BN dataset.}
\label{tab:dataset-summary}
\small
\setlength{\tabcolsep}{4.5pt}
\renewcommand{\arraystretch}{1.14}
\resizebox{\columnwidth}{!}{%
\begin{tabular}{@{}lr@{}}
\toprule
\textbf{Dataset Attribute} & \textbf{Specification / Count} \\
\midrule
\multicolumn{2}{@{}l}{\textit{\textbf{Corpus Scale \& Partitions}}} \\
\quad Total structured tutoring traces & 4,099 \\
\quad Training split ($D_{\text{train}}$) & 3,667 (89.5\%) \\
\quad Held-out evaluation split ($D_{\text{eval}}$) & 432 (10.5\%) \\
\midrule
\multicolumn{2}{@{}l}{\textit{\textbf{Curriculum \& Pedagogical Scope}}} \\
\quad Target learner level & CEFR A1--A2 (Beginner) \\
\quad NCTB curriculum grammar units & 13 units \\
\quad Thematic topic clusters & 15 situational domains \\
\midrule
\multicolumn{2}{@{}l}{\textit{\textbf{Trace Schema \& Sequence Lengths}}} \\
\quad Structured fields per trace & 7 multi-task fields \\
\quad Mean Bangla source length & 5.5 words \\
\quad Mean English natural translation length & 6.6 words \\
\bottomrule
\end{tabular}%
}
\end{table}

\section{Model Adaptation and Evaluation}
\label{sec:method}

\subsection{LoRA Fine-Tuning}

\begin{figure}[!t]
\centering
\includegraphics[width=0.74\columnwidth]{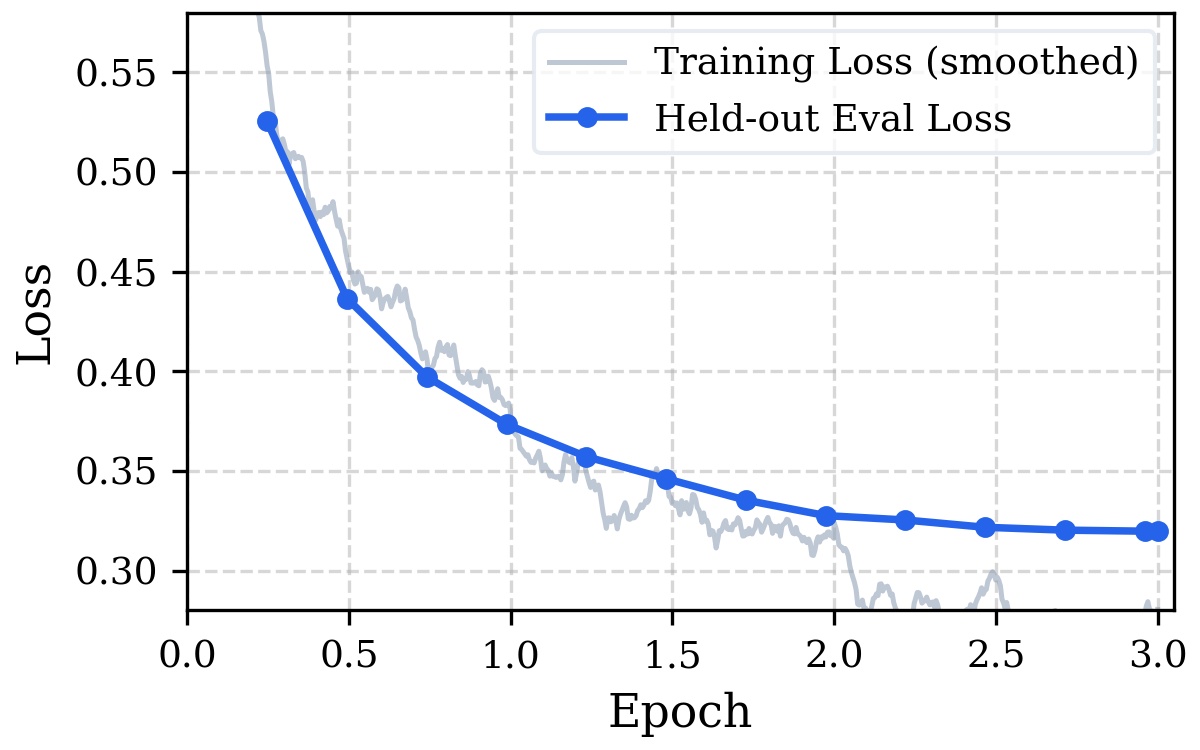}
\vspace{-6pt}
\caption{Training loss progression of Qwen3-0.6B on TRACE-BN across three epochs (621 steps). Loss decreases steadily and stabilizes near training completion.}
\label{fig:loss_curve}
\vspace{-8pt}
\end{figure}

We fine-tune Qwen3-0.6B~\cite{team2025qwen3} using LoRA~\cite{hu2021lora} with 4-bit quantization through Unsloth on a single 16\,GB Google Colab T4 GPU. The training input is the Bangla sentence together with the target schema specification; the model is trained to generate the complete seven-field tutoring trace as a single response. LoRA uses rank $r=16$, scaling factor $\alpha=32$, zero dropout, and targets the \texttt{q, k, v, o, gate, up, down} projection layers. Training runs for a fixed schedule of three epochs (621 optimizer steps) with a cosine learning-rate schedule peaking at $2\times10^{-4}$; loss is logged at 51-step intervals (Fig.~\ref{fig:loss_curve}). The final-step checkpoint is evaluated on the 432-example split without hyperparameter tuning or checkpoint cherry-picking on evaluation data.

\subsection{Baseline Models}

We compare the tuned Qwen3-0.6B with its untuned base model and three zero-shot baselines: Gemma-4 E2B (2.3B) as a larger-model reference, TigerLLM-1B as a Bangla-specialized baseline, and Llama-3.2-1B-Instruct as a general multilingual baseline. All models receive the same task instruction, formatted through each model family's native chat template, and are evaluated on the same 432 held-out Bangla inputs; the comparison models are not fine-tuned. We also piloted BanglaLlama-3.2-3B on a stratified sample of 18 inputs but excluded it from the full comparison after it produced 0\% schema-valid outputs on that sample; the model could not complete the task in a form our metrics could score.
\begin{figure*}[t]
\centering
\includegraphics[width=0.80\textwidth]{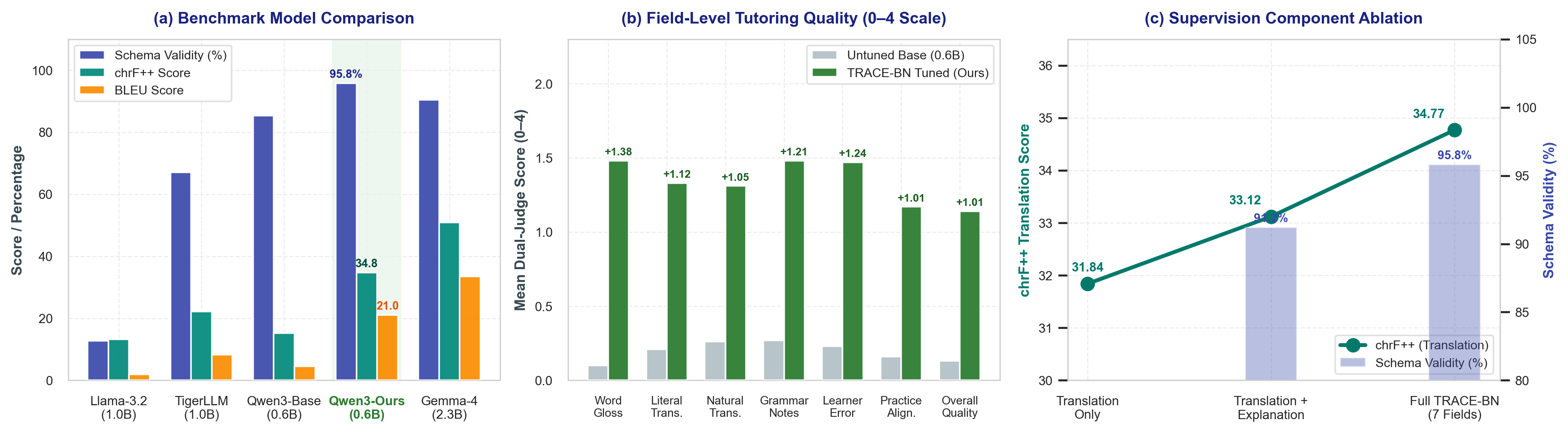}
\caption{Empirical evaluation of TRACE-BN adaptation: (a) Benchmark model comparison against zero-shot baselines on schema validity (\%), chrF++, and BLEU; (b) Field-level pedagogical ratings on a 0--4 anchored scale evaluated independently by dual automated judges (GPT-4o-mini and DeepSeek-V4-Flash-0731); (c) Trace component ablation demonstrating the impact of multi-task supervision on translation and schema adherence.}
\label{fig:performance_chart}
\end{figure*}

\subsection{Evaluation}

Schema validity requires a response to be valid JSON with all seven fields present, including a parseable English \texttt{natural\_translation} field. For BLEU~\cite{papineni2002bleu} and chrF++~\cite{popovic2017chrf}, scores are computed on extracted \texttt{natural\_translation} strings when parseable; outputs lacking a valid translation field are scored as empty strings against the reference. Because the teacher model generates both training traces and silver references, these metrics measure agreement with the teacher reference rather than independently adjudicated translation quality.

To obtain a more granular assessment of tutoring behavior, we conduct an additional field-level evaluation on the held-out split using two independent automated judges, GPT-4o-mini and DeepSeek-V4-Flash-0731. Each judge receives the Bangla source sentence, the corresponding TRACE-BN reference trace, and the generated response. It then rates seven dimensions (word-level gloss accuracy, literal translation accuracy, natural translation accuracy, grammar explanation accuracy, learner-error diagnosis, practice alignment, and overall tutoring quality) on a 0--4 anchored ordinal scale, where 4 denotes fully correct and 0 denotes unusable output. The reference trace provides the intended instructional target against which the generated response is scored.

The two judges score the same 432 held-out outputs. We report mean field-level scores, the proportion of outputs receiving a score of at least 3, and inter-judge agreement via quadratic weighted Cohen's $\kappa$~\cite{zheng2023judging}. This field-level analysis distinguishes local errors from failures affecting the tutoring response more broadly.

For metric comparisons, we use paired bootstrap resampling with 1,000 iterations and report $p$-values for the main comparisons.

\section{Results}
\label{sec:results}

We first compare the tuned model with the untuned and zero-shot baselines, then examine pedagogical quality and failure modes, the contribution of individual trace components, qualitative behavior, and structural reliability.

\subsection{Overall Performance}

\begin{table}[t]
\centering
\caption{Performance of zero-shot baselines and the TRACE-BN-tuned Qwen3-0.6B on the shared held-out set ($n=432$). Schema validity requires valid JSON with all seven fields present, including a parseable English \texttt{natural\_translation}; translation metrics use the teacher-model output as a silver reference; Tok/s is inference throughput on the evaluation set.}
\label{tab:main-results}
\small
\setlength{\tabcolsep}{4pt}
\renewcommand{\arraystretch}{0.96}
\resizebox{\linewidth}{!}{%
\begin{tabular}{@{}lccccc@{}}
\toprule
\textbf{Model} & \textbf{Params} & \textbf{Schema} & \textbf{chrF++} $\uparrow$ & \textbf{BLEU} $\uparrow$ & \textbf{Tok/s} \\
\midrule
Gemma-4 E2B & 2.3B & 90.5\% & 50.91 & 33.54 & 45.97 \\
TigerLLM-1B & 1.0B & 67.1\% & 22.23 & 8.21 & 48.79 \\
Llama-3.2-1B-Instruct & 1.0B & 12.7\% & 13.21 & 1.95 & 73.11 \\
Qwen3-0.6B (base) & 0.6B & 85.4\% & 15.28 & 4.52 & 90.89 \\
\midrule
\textbf{Qwen3-0.6B (ours)} & \textbf{0.6B} & \textbf{95.8\%} & \textbf{34.77} & \textbf{21.03} & \textbf{75.47} \\
\bottomrule
\end{tabular}%
}
\vspace{1pt}
\begin{flushleft}
\footnotesize
Ours improves over the untuned Qwen3-0.6B by $+19.49$ chrF++ (paired bootstrap, 1,000 resamples, $p<0.05$).
\end{flushleft}
\vspace{-4pt}
\end{table}

The tuned Qwen3-0.6B improves both structural validity and reference-based translation scores over the untuned model (Fig.~\ref{fig:performance_chart}(a), Table~\ref{tab:main-results}). Schema validity increases from 85.4\% to 95.8\%, while chrF++ and BLEU rise from 15.28 to 34.77 and from 4.52 to 21.03, respectively. The chrF++ gain over the base model is $+19.49$ ($p<0.05$, paired bootstrap), while the gain over TigerLLM-1B is $+12.54$ ($p<0.05$). Gemma-4 E2B remains the strongest translation baseline, with 50.91 chrF++ and 33.54 BLEU. On inference throughput, the tuned model reaches 75.47 tokens/s, comparable to the untuned base model (90.89 tokens/s) and above Gemma-4 E2B (45.97 tokens/s) despite Gemma-4 E2B's 2.3B parameters, consistent with the offline, on-device deployment the model targets.

\subsection{Pedagogical Quality and Failure Analysis}

\begin{table}[t]
\centering
\caption{Field-level pedagogical evaluation of the base and TRACE-BN-tuned Qwen3-0.6B on the shared held-out set ($n=432$). Scores are means on a 0--4 scale; the percentage in parentheses denotes outputs receiving a score of at least 3. Inter-judge agreement is reported using quadratic weighted Cohen's $\kappa$.}
\label{tab:field-pedagogical}
\small
\setlength{\tabcolsep}{3.5pt}
\renewcommand{\arraystretch}{0.96}
\resizebox{\columnwidth}{!}{%
\begin{tabular}{@{}lccc@{}}
\toprule
\textbf{Dimension} &
\textbf{Base} &
\textbf{Ours} &
\textbf{$\kappa$} \\
\midrule
Word gloss accuracy
& 0.10\;(0.0\%)
& \textbf{1.48}\;(16.4\%)
& 0.788 \\
Literal translation
& 0.21\;(0.0\%)
& \textbf{1.33}\;(14.8\%)
& 0.754 \\
Natural translation
& 0.26\;(1.9\%)
& \textbf{1.31}\;(16.2\%)
& 0.803 \\
Grammar explanation
& 0.27\;(0.0\%)
& \textbf{1.48}\;(9.7\%)
& 0.489 \\
Learner-error diagnosis
& 0.23\;(0.0\%)
& \textbf{1.47}\;(11.3\%)
& 0.585 \\
Practice alignment
& 0.16\;(0.2\%)
& \textbf{1.17}\;(9.7\%)
& 0.529 \\
Overall tutoring quality
& 0.13\;(0.0\%)
& \textbf{1.14}\;(7.5\%)
& 0.700 \\
\bottomrule
\end{tabular}%
}
\vspace{-2pt}
\end{table}

Schema validity alone does not establish tutoring quality. The field-level evaluation shows that TRACE-BN fine-tuning improves every evaluated tutoring dimension relative to the untuned Qwen3-0.6B (Fig.~\ref{fig:performance_chart}(b), Table~\ref{tab:field-pedagogical}). The largest gains are observed in word-level gloss accuracy ($+1.38$), learner-error diagnosis ($+1.24$), grammar explanation accuracy ($+1.21$), and literal translation accuracy ($+1.12$). The tuned model also improves natural translation quality from 0.26 to 1.31 and overall tutoring quality from 0.13 to 1.14 on the 0--4 scale. Although the absolute scores remain below the upper end of the rubric, the consistent improvement across all fields suggests that the effect of fine-tuning extends beyond output formatting.

Inter-judge agreement is substantial for word-level glosses ($\kappa=0.788$), literal translation ($\kappa=0.754$), natural translation ($\kappa=0.803$), and overall tutoring quality ($\kappa=0.700$), with moderate agreement for grammar explanation ($\kappa=0.489$), learner-error diagnosis ($\kappa=0.585$), and practice alignment ($\kappa=0.529$). These differences are expected because fine-grained pedagogical judgments involve greater interpretive variation than translation correctness. Overall, the agreement results support the use of field-level automated evaluation as a scalable complement to human validation rather than as a substitute for learner studies.

Qualitative inspection indicates that remaining errors are concentrated mainly in held-out idiomatic expressions and in fine-grained word-level gloss decisions. These localized failures are distinct from the structural failures captured by schema validity.

\subsection{Trace Component Ablation}

\begin{table}[t]
\centering
\caption{Ablation of tutoring trace components on Qwen3-0.6B ($n=432$).}
\label{tab:ablation}
\small
\setlength{\tabcolsep}{5pt}
\renewcommand{\arraystretch}{0.96}
\resizebox{\columnwidth}{!}{%
\begin{tabular}{@{}lccc@{}}
\toprule
\textbf{Training Configuration} & \textbf{Target Fields} & \textbf{Schema} & \textbf{chrF++} $\uparrow$ \\
\midrule
Translation Only & 1 (String) & N/A & 31.84 \\
Translation + Explanation & 2 (JSON) & 91.2\% & 33.12 \\
\textbf{Full TRACE-BN (Ours)} & \textbf{7 (JSON)} & \textbf{95.8\%} & \textbf{34.77} \\
\bottomrule
\end{tabular}%
}
\vspace{-2pt}
\end{table}

We evaluate three supervision configurations (Fig.~\ref{fig:performance_chart}(c), Table~\ref{tab:ablation}). Translation-only training reaches 31.84 chrF++, while adding Bangla explanations increases chrF++ to 33.12. The full seven-field TRACE-BN configuration reaches 34.77 chrF++, suggesting that richer structured supervision benefits even the translation component alone, rather than only adding auxiliary output fields.

\subsection{Qualitative Behavior}

Beyond schema compliance, the tuned model generates fluent translations with targeted grammar notes (Fig.~\ref{fig:qualitative}). For question formation, it correctly identifies auxiliary \textit{do} insertion rather than translating the Bangla question marker literally.

\subsection{Structured Reliability}

The 95.8\% schema validity rate corresponds to 18 invalid outputs ($n=432$), versus 63 for the base model. To confirm these results transfer to the target deployment setting, we additionally exported the merged adapter to GGUF format and confirmed matching outputs under \texttt{llama.cpp} local inference.

\section{Conclusion}

We presented TRACE-BN, a curriculum-guided dataset for structured Bangla-to-English tutoring, and showed that Qwen3-0.6B can learn to generate seven-field tutoring traces with 95.8\% schema validity and a $+19.49$ chrF++ improvement over the base model. Field-level evaluation with two independent judges shows consistent improvement across all evaluated tutoring dimensions, with moderate-to-substantial inter-judge agreement, while a 100-trace human audit supports the quality of the supervision data. These results show that curriculum-guided structured supervision can transfer tutoring structure and content to a sub-1B model, achieving higher schema validity and throughput than the larger Gemma-4 E2B zero-shot baseline, while remaining behind it on translation quality. The same supervision strategy provides a basis for future extensions to other language pairs and curricula; establishing whether such tutors improve learners' English proficiency will require controlled learner studies.

\section*{AI Disclosure}
LLMs were used for teacher-trace synthesis and editorial assistance; human authors designed all methodology and verified all results.

\bibliographystyle{IEEEtran}
\bibliography{references}

\end{document}